\documentclass[10pt,twocolumn,letterpaper]{article}

\usepackage{cvpr}
\usepackage{times}
\usepackage{epsfig}
\usepackage{graphicx}
\usepackage{amsmath}
\usepackage{amssymb}
\usepackage{subfigure}
\usepackage[marginal]{footmisc}

\usepackage[pagebackref=true,breaklinks=true,letterpaper=true,colorlinks,bookmarks=false]{hyperref}

\cvprfinalcopy 

\def\cvprPaperID{****} 

\ifcvprfinal\fi

\begin{document}

\title{Graph based Dynamic Segmentation of Generic Objects in 3D\footnotemark}

\author{Xiao Lin, Josep R. Casas and Montse Pard\`{a}s\\
Technical University of Catalonia (UPC)\\
Jordi Girona 1-3, 08034 Barcelona\\
{\tt\small james.linxiao@gmail.com}
}


\maketitle

\footnote[1]{*This work has been developed in the framework of the project TEC2013-43935-R, financed by the Spanish Ministerio de Economía y Competitividad and the European Regional Development Fund (ERDF)}
\section{Introduction}
3D segmentation is a promising building block for high level applications such as scene understanding and interaction analysis. New challenges emerge for computer vision techniques in generic scenarios with RGBD stream data. We focus on temporally evolving 3D point clouds in order to segment an image sequence into regions, which should {\em ideally} correspond to meaningful objects in the scene. To achieve this goal, some approaches incorporate high level knowledge into the segmentation process, such as object models~\cite{felzenszwalb2010object} and accurate object annotations in the initialization stage~\cite{tsai2012motion}. However, most computer vision applications involve large amounts of data with different types of scenes containing several objects, which makes those methods difficult to be adapted to generic scenes.

In order to segment generic objects in the 3D scene, without explicit object models or accurate initialization, we will consider that objects correspond to "compact point clouds" in the 3D-space plus time domain. However, point clouds corresponding to an object can break into different connected components due to occlusions, or can merge with point clouds corresponding to other objects, producing a single connected component, when they become spatially close (object interaction). Our system produces a robust spatio-temporal segmentation of the point clouds, analyzing their connectivity to define the objects according to the evidence observed up to a given temporal point.

To tackle spatio-temporal connectivity 
Hickson et al.\cite{2014-Hickson-EHGSRV} propose a graph based model to hierarchically perform a video segmentation from over-segmented frames. But the over-segmentation in different frames is calculated independently, which may lead to a temporal consistency problem. Abramov et al.\cite{abramov2012depth} propose to transfer labels from frame to frame, relying on optical flow. But this requires a good initialization and is restricted to the performance of the optical flow estimation technique. Husain et al. \cite{husain2015consistent} maintain a quadratic surface model to generally represent the object segments in the scene. But it is difficult to handle objects with large displacements in successive frames.

We propose a novel 3D segmentation method for RGBD stream data to deal with the above mentioned problems in a generic scenario with frequent object interactions. It mainly contributes in two aspects, while being generic and not requiring initialization: firstly, a novel tree structure representation for the point cloud of the scene is proposed. Then, a dynamic management mechanism for connected component splits/merges exploits the tree structure representation. Experiments illustrate promising results and its potential application in object segmentation and interaction analysis.
\section{Graph based dynamic 3D segmentation}
\begin{figure}
    \centering
    \includegraphics[width=0.9\linewidth,height=1.3in]{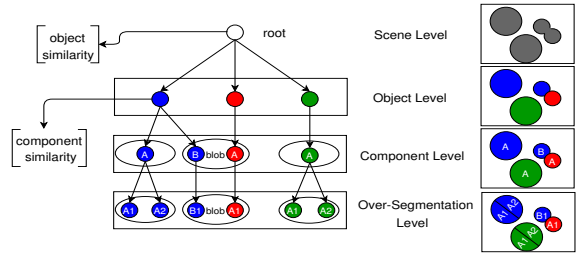}
    \caption{Tree structure representation. Left: an example of the tree. Right: point cloud views in different levels of the tree}
    \label{fig:tree_structure}
\end{figure}

\subsection{Tree structure representation of the point cloud}\label{tree structure definition}
\begin{figure}
    \centering
    \includegraphics[width=1.0\linewidth,height=0.9in]{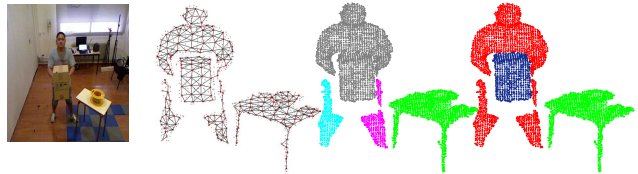}
    \caption{Segmentation example. Left to right: color image, point cloud graph, blobs and objects obtained from our segmentation}
    \label{fig:segmentation}
\end{figure}
We represent the foreground point cloud as a graph $G$. The set of connected nodes of this graph define the connected components of the point cloud, which we call blobs. As shown in Fig.\ref{fig:tree_structure}, the root of the tree represents this graph G. At the second level, we represent the objects in the scene. The next level, named component level, is used to handle possible splits and merges of objects. An object is represented by more than one components if it splits in different blobs. Components belonging to different objects can be part of the same blob, because of the interactions between objects. The next level of the tree, is the over-segmentation level. We over-segment components into segments in order to correctly establish correspondences between trees along time and update its structure, that is, the temporal coherent assignment of labels to the segmented objects.
\subsection{Tree structure creation}
Taking the point cloud in frame $t$ as input data, we abstract it with super voxels, using the method proposed in \cite{papon2013voxel}. Then, a graph $G$ is constructed regarding the spatial connectivity between super voxels. We group the point cloud into blobs by detecting the connected components in the graph.

The tree in the first frame is created by simply taking the detected blobs as the objects. Accordingly, we create one component for each object and over-segment each component into segments. Apart from the first frame, the tree is built in a bottom-up way, starting at the component level. First, a correspondence is made between the connected components of the graph (blobs) and the segments of the over-segmentation level in the previous frames. The over-segmentation level is employed to avoid temporal consistency problem (one component in the previous frame may split in different blobs in the current frame). Establishing the correspondence between the blob labels and the segments is a problem of assigning $M_{b}$ blob labels to $M_{s}$ segments. This is a nonlinear integer programming problem which is solved using a Genetic Algorithm to minimize an energy function which is composed of two terms: one for representing the appearance changes and one for the displacements.
A further segmentation is needed when segments that correspond to different objects in the previous frame are assigned to the same blob. A restricted graph cut method is employed to segment the graph of the blob by minimizing a segmentation energy function, in which we consider the degree that a graph cut fits the current data while being coherent with the minimum cut in the previous frame.
Once the current segmentation is done, the components and objects in the current tree are created initially from it regarding the previous tree structure.

To dynamically manage object splits and merges, we maintain similarities between nodes at the component and object level respectively and update the tree based on it. The component similarities are measured among components which belong to the same object while the object similarities are measured among objects. These similarities are computed considering spatial distance and color difference, which reveal the likelihood of object splits and merges. We accumulate them along time by averaging the current similarity and the previous accumulated similarity regarding the established correspondences. Then object splits and merges are confirmed by thresholding the accumulated similarities.

Finally, an over-segmentation is performed at the component level to generate segments for establishing the correspondence to the next frame. Specifically, a normalized cut is performed on the graph representing the component iteratively until the cut cost is larger than a threshold $T_{o}$.
\section{Evaluation}
\begin{figure}
\begin{center}
    \subfigure[]{
    \includegraphics[width=0.31\linewidth,height=0.55in]{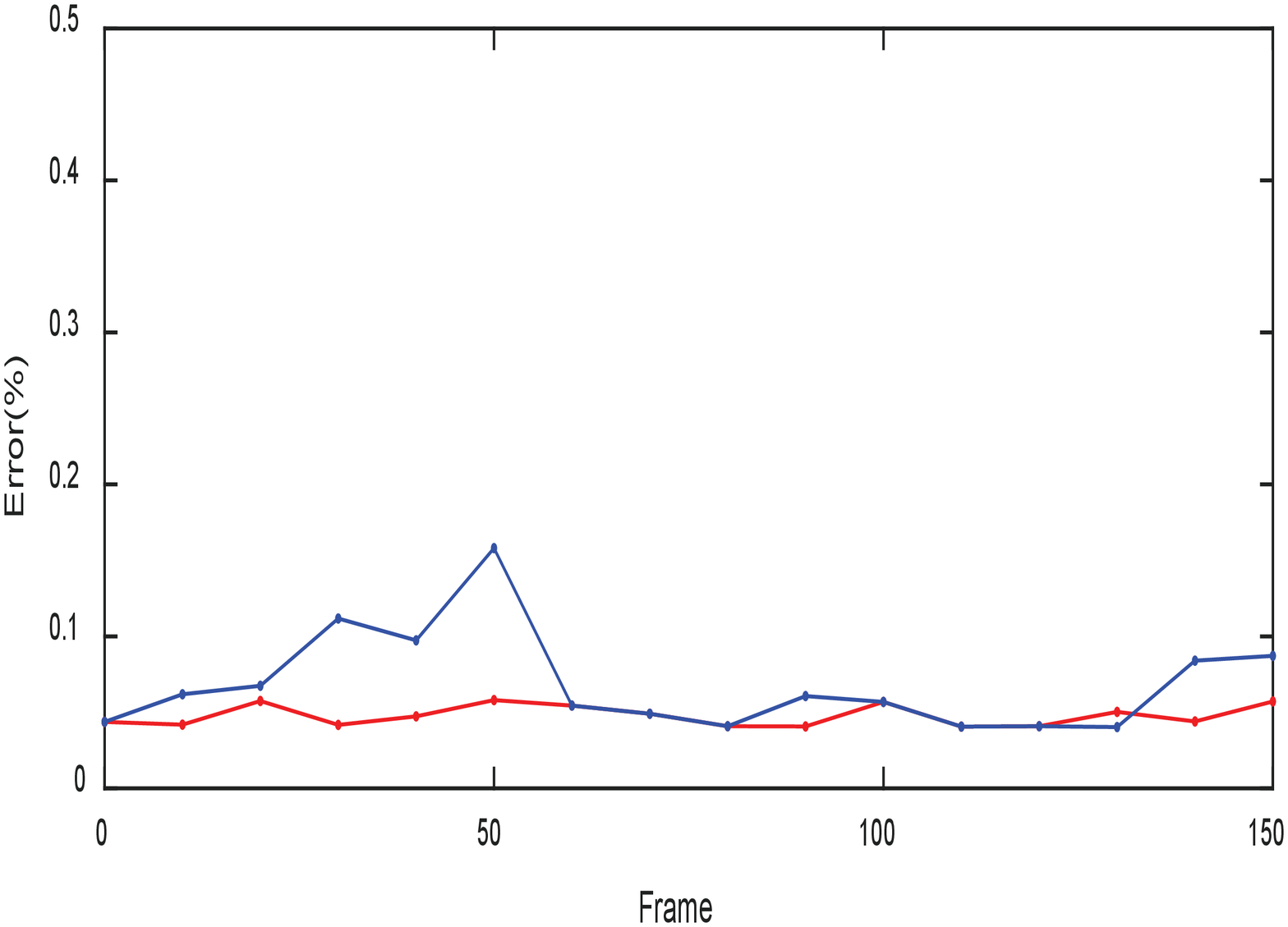}}
    \subfigure[]{
    \includegraphics[width=0.31\linewidth,height=0.55in]{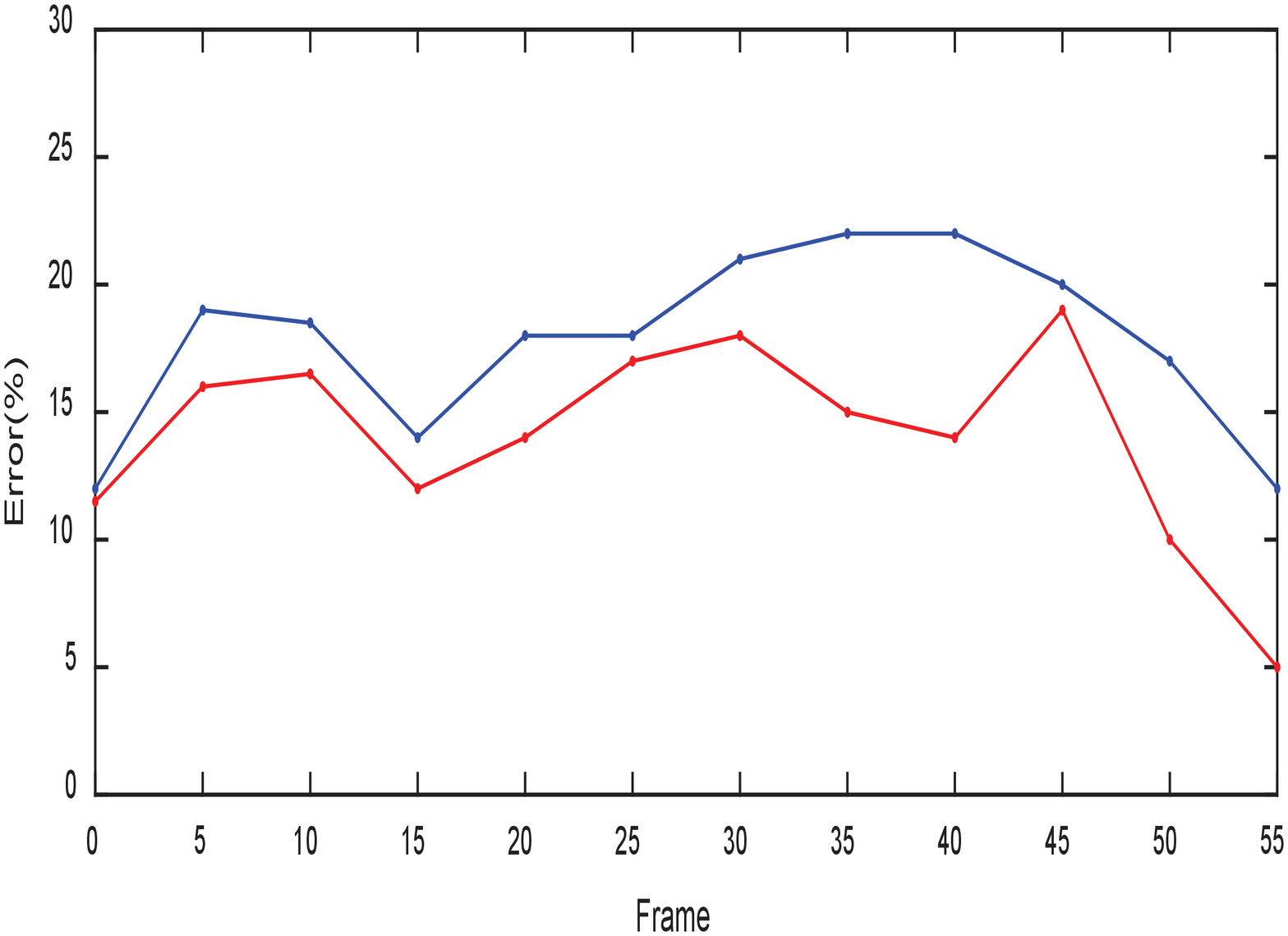}}
    \subfigure[]{
    \includegraphics[width=0.31\linewidth,height=0.55in]{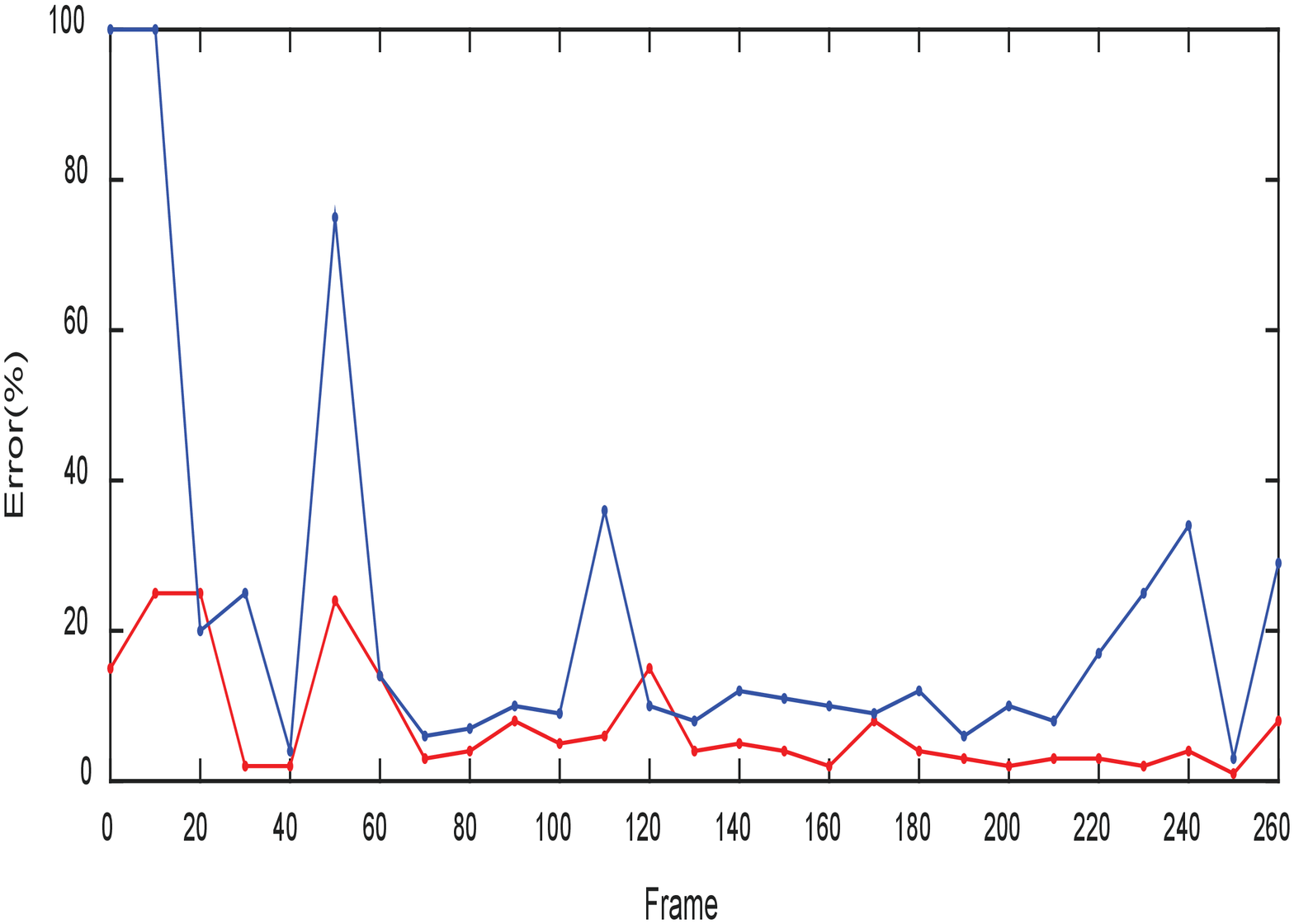}}
\end{center}
\caption[]{\centering Quantitative error of GDS (in red) and the method in \cite{husain2015consistent} (in blue) for the 3 sequences provided in \cite{husain2015consistent}.}
\label{fig:comparison_quantitative}
\end{figure}

Our approach is capable to segment objects from RGBD stream data without initialization or explicit model and also obtain the interactions between objects, which is implicit in the tree structure. An interaction is made when a blob is related to more than one object label. Experiments on a human manipulation data set~\cite{pieropan2014audio} with 3D ground truth labeling are designed to qualitatively and quantitatively evaluate the proposed Graph based Dynamic Segmentation approach (GDS). GDS achieves an overall $3.92\%$ mean segmentation error and detects $693$ out of $788$ interactions in this data set. For comparison, we employ 3 sequences provided in \cite{husain2015consistent}. Fig.\ref{fig:comparison_quantitative} shows the better quantitative segmentation results from GDS comparing to the method in \cite{husain2015consistent}. Visual experiment results are available on https://imatge.upc.edu/web/node/1806.

{\scriptsize
\bibliographystyle{ieee}
\bibliography{egbib}
}

\end{document}